\documentclass{article}
\usepackage{spconf}
\usepackage{cite}
\usepackage{amsmath,amssymb,amsfonts}
\usepackage{graphicx}
\usepackage{textcomp}
\usepackage{lipsum, color}
\usepackage{graphics}
\usepackage{titlesec}
\usepackage{subfig}
\usepackage{graphicx}
\usepackage{xcolor}
\usepackage{titlesec}
\usepackage{subfig}
\usepackage{graphicx}
\usepackage{mathtools}
\usepackage{siunitx}
\usepackage{soul}
\usepackage{balance}
\usepackage{algpseudocode}
\usepackage{cite}
\usepackage[font=footnotesize]{caption}
\usepackage{tabularx}
\usepackage{array}
\usepackage{floatrow}
\usepackage[english]{babel}
\usepackage[utf8x]{inputenc}
\usepackage{amsmath}
\usepackage{xcolor}
\usepackage{tikz}
\usepackage{blindtext}
\usepackage{enumitem}
\usepackage[T1]{fontenc}
\usepackage{flushend}

\def\BibTeX{{\rm B\kern-.05em{\sc i\kern-.025em b}\kern-.08em
    T\kern-.1667em\lower.7ex\hbox{E}\kern-.125emX}}

\begin{document}
\title{Robust Monocular Localization of Drones by Adapting Domain Maps to Depth Prediction Inaccuracies}

\name{Priyesh Shukla, Sureshkumar S., Alex C. Stutts, Sathya Ravi, Theja Tulabandhula, and Amit R. Trivedi}

\address{University of Illinois at Chicago, USA}
\maketitle

\begin{abstract}
We present a novel monocular localization framework by jointly training deep learning-based depth prediction and Bayesian filtering-based pose reasoning. The proposed cross-modal framework significantly outperforms deep learning-only predictions with respect to model scalability and tolerance to environmental variations. Specifically, we show little-to-no degradation of pose accuracy even with extremely poor depth estimates from a lightweight depth predictor. Our framework also maintains high pose accuracy in extreme lighting variations compared to standard deep learning, even without explicit domain adaptation. By openly representing the map and intermediate feature maps (such as depth estimates), our framework also allows for faster updates and reusing intermediate predictions for other tasks, such as obstacle avoidance, resulting in much higher resource efficiency.  
\end{abstract}
\begin{keywords}
Depth neural network, drone localization.
\end{keywords}

\vspace{-5pt}
\section{Introduction}
For self-navigation, the most fundamental computation required for a vehicle is to determine its position and orientation, i.e., \textit{pose} during motion. Higher-level path planning objectives such as motion tracking and obstacle avoidance operate by continuously estimating vehicle's pose. Recently, deep neural networks (DNNs) have shown a remarkable ability for vision-based pose estimation in highly complex and cluttered environments \cite{kendall2015posenet, zhou2020kfnet, wang2020atloc}. For visual pose estimation, DNNs can learn the correlation of vehicle's position/orientation and visual fields to a mounted camera. Thereby, vehicle's pose can be predicted using a monocular camera alone. In contrast, the traditional methods required bulky and power-hungry range sensors or stereo vision sensors to resolve the ambiguity between an object's distance and its scale \cite{fox2001particle,skrzypczynski2017mobile}.

However, DNN's \textit{implicit learning} of flying domain features such as its map, placement of objects, coordinate frame, domain structure, \textit{etc.} in a standard pose-DNN also affects the robustness and adaptability of pose estimations. The traditional filtering-based approaches \cite{thrun2002probabilistic} account for the flying space structure using explicit representations such as voxel grids, occupancy grid, Gaussian mixture model (GMM), \textit{etc.} \cite{Dhawale-2020-121381}; thereby, updates to the flying space such as map extension, new objects, and locations can be more easily accommodated. Comparatively, DNN-based estimators cannot handle selective map updates, and the entire model must be retrained even under small randomized or structured perturbations. Additionally, filtering loops in traditional methods can adjudicate predictive uncertainties against measurements to systematically prune hypothesis space and can express prediction confidence along with the prediction itself \cite{thrun2002particle}. Whereas feedforward pose estimations from a deterministic DNN are vulnerable to measurement and modeling uncertainties.

\begin{figure*}[t]
  \centering
  \includegraphics[width=0.85\textwidth]{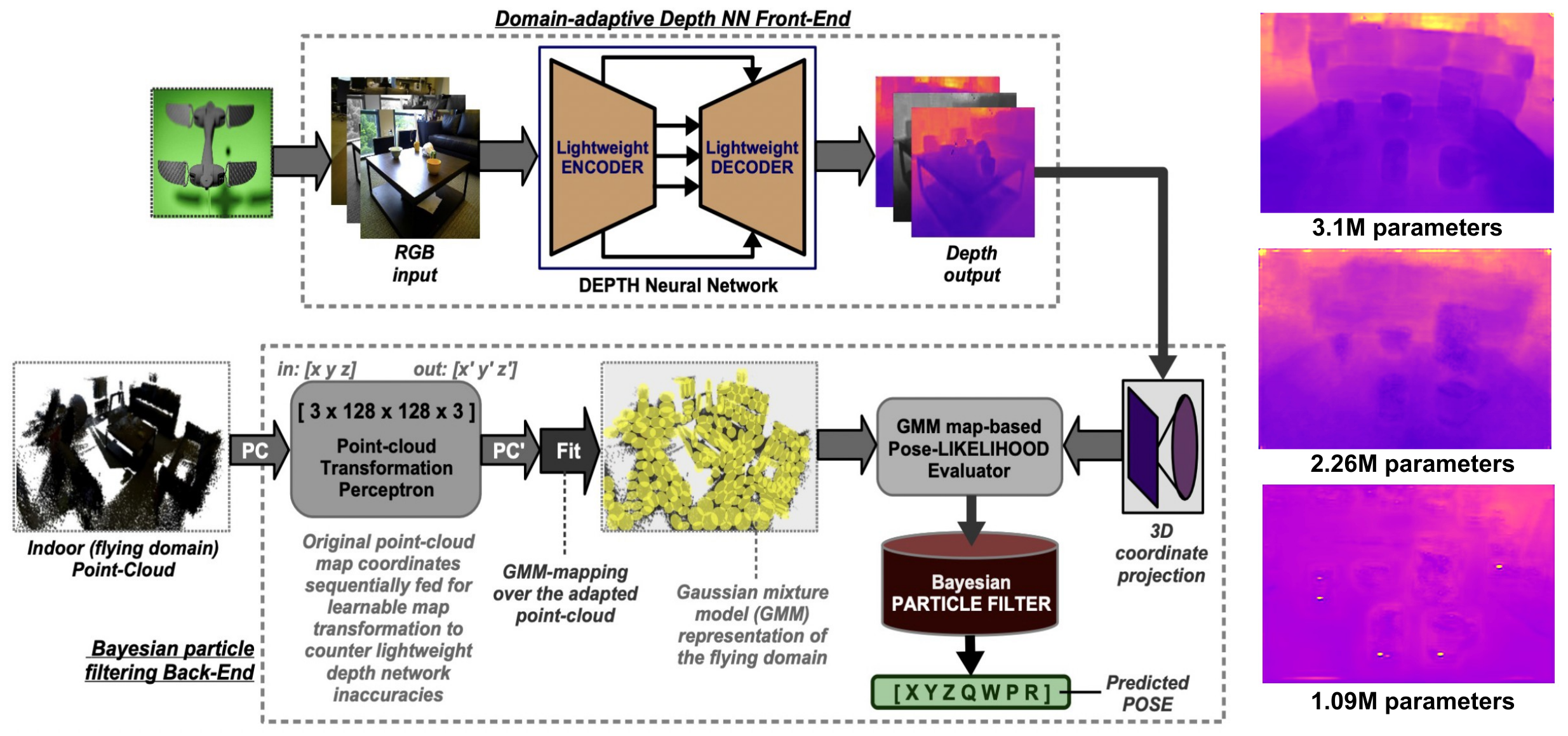}
  \caption{Proposed framework integrating depth estimator front-end and particle filter back-end for extremely lightweight and robust localization. DNN-based preprocessing avoids area/power-hungry range sensors. The filtered response of the back-end predictor is robust against measurement and modeling uncertainties. On the right are depth predictions from a lightweight network with varying model sizes.}
  \label{fig:mainfig}
  \vspace{-10pt}
\end{figure*}

In thie paper, we use integrate traditional filtering techniques with deep learning to overcome such limitations of DNN-based pose estimation while exploiting their suitability to operate efficiently with monocular cameras alone. Specifically, we present a novel framework for visual localization by integrating DNN-based depth prediction and Bayesian filtering-based pose localization. In Figure \ref{fig:mainfig}, avoiding range sensors for localization, we utilize a DNN-based lightweight depth prediction network at the front end and sequential Bayesian estimation at the back end. Our key observation is that, unlike pose estimation, which innately depends on map characteristics such as spatial structure, objects, coordinate frame, \textit{etc.}, depth prediction is {\em map-independent} \cite{godard2017unsupervised, wofk2019fastdepth}. Thus, by applying deep learning only on domain-independent tasks and utilizing traditional models where domain is openly (or explicitly) represented helps improve the predictive robustness. Limiting deep learning to only domain-independent tasks also allows our framework to utilize vast training sets from unrelated domains. Open representation of map and depth estimates enables faster domain-specific updates and utilization of intermediate feature maps for other autonomy objectives, such as obstacle avoidance, thus improving computational efficiency.

\section{Monocular Localization with Depth Neural Network and Pose Filters}
In Figure \ref{fig:mainfig}, our framework integrates deep learning-based depth prediction and Bayesian filters for visual pose localization in the 3D space. At the front end, a depth DNN scans monocular camera images to predict the relative depth of image pixels from the camera's focal point. A particle filter localizes the camera pose at the back end by evaluating the likelihood of 3D projection of depth scans over a GMM-based map representation of 3D space. Both frameworks are jointly trained for the extremely lightweight operation. Various components of the framework are discussed below: 
\vspace{-5pt}

\subsection{Extremely Lightweight Depth Prediction}
DNN-based monocular depth estimation has gained wide interest owing to impressive results. Several fully supervised \cite{zioulis2018omnidepth}, self-supervised \cite{godard2019digging}, and semi-supervised \cite{guizilini2020robust} convolutional neural network (CNN)-based depth estimators have been presented with promising results. However, for low-power edge robotics \cite{floreano2015science}, the existing depth DNNs are often oversized. A typical depth DNN combines an encoder that extracts the relevant features from the input images. The features are then up-sampled using a decoder to predict the depth map. \textit{Skip connections} between various encoding and decoding layers are typically used to obtain high-resolution image features within the encoder which in-turn helps the decoding layers reconstruct a high resolution depth output. 

\begin{figure}[h]
  \centering
  \includegraphics[width=\textwidth]{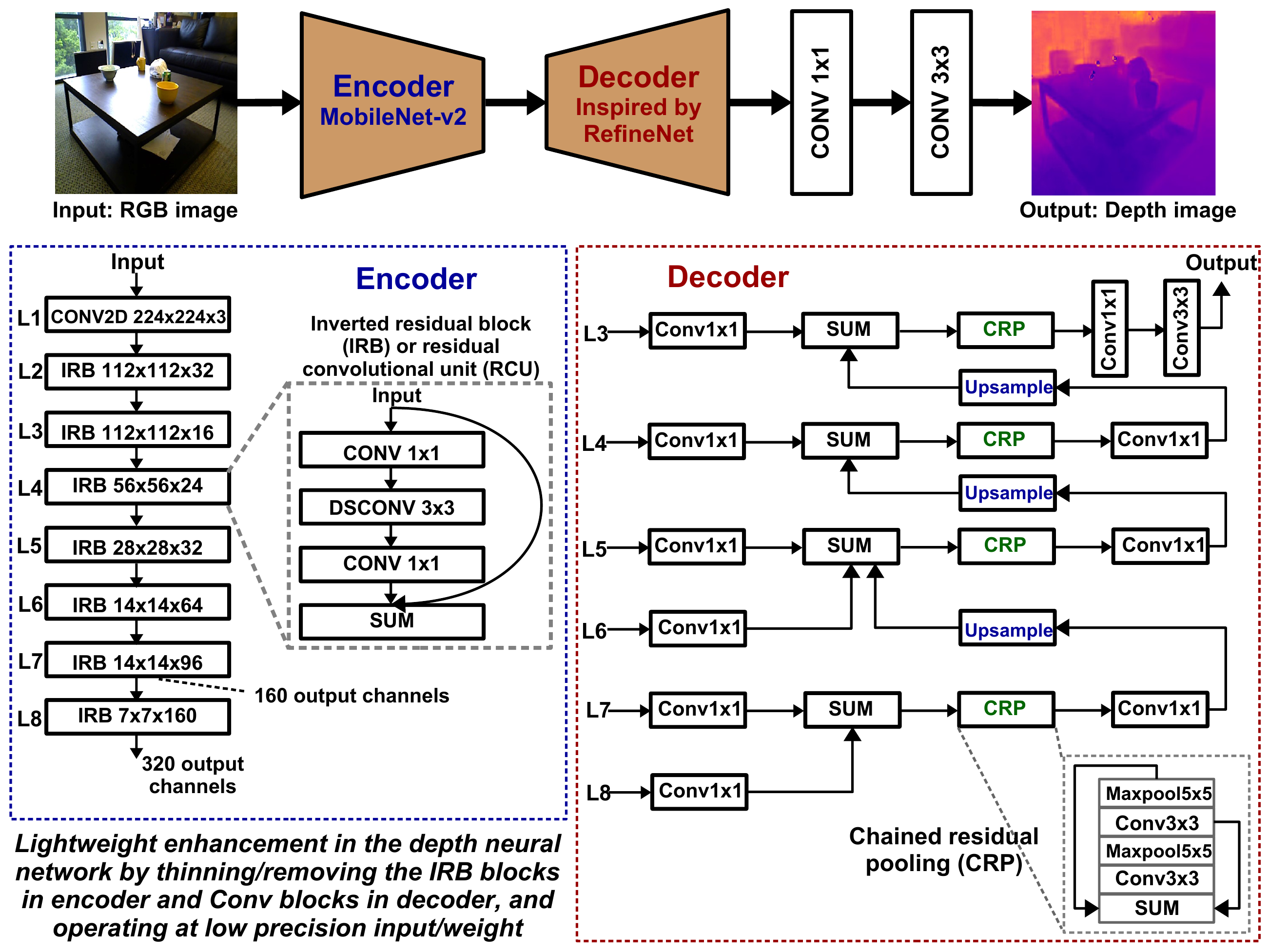}
  \caption{Depth neural network architecture with MobileNet-v2 \cite{sandler2018mobilenetv2} encoder and decoder based on RefineNet \cite{nekrasov2019real}.}
  \vspace{-0.5em}
  \label{fig:depthNN}
\end{figure}

In Figure \ref{fig:depthNN}, we consider a depth DNN that integrates state-of-the-art architectures for lightweight processing on mobile devices. The depth predictor uses MobileNet-v2 as encoder and RefineNet \cite{nekrasov2019real} as decoder. MobileNet-v2 concatenates memory-efficient inverted residual blocks (IRBs). The intermediate layer outputs (or RGB-image features) from the encoder are decoded through the successive channels of convolutional sum, chained residual pooling (CRP), and depth-feature upsampling. This architecture uniquely utilizes only 1$\times$1 convolutional layers in SUM and CRP blocks (replacing traditional high receptive field 3$\times$3 CONV layers with 1$\times$1 CONV layers), thus significantly reducing model parameters. Due to the modular architecture of the depth predictor in Figure \ref{fig:depthNN}, its size can be scaled down by reducing the number of layers in the encoder and decoder. However, with fewer parameters, the prediction quality is affected. Figure \ref{fig:mainfig} (on the right) shows the depth quality by reducing the number of model parameters. Later, we will discuss how despite lower quality depth prediction, accurate pose localization can be achieved by adapting maps to depth inaccuracies.   

\vspace{-20pt}
\subsection{Memory Efficient Mapping using GMMs}
To minimize the memory footprint of maps, we utilized a GMM-based representation of 3D maps \cite{dhawale2018fast}. The point-cloud distribution of tested maps was clustered and fitted with a 3D GMM using Expectation-Maximization (EM) procedures. Although alternate map representations are prevalent, the parametric formulation of GMMs can considerably minimize the necessary storage and extraction cost. For example, Voxel grids \cite{moravec1985high} use cells to represent dense maps and are simpler to extract. However, the representation suffers from storage inefficiency since the free space also needs to be encoded. Surfels, 3D planes, and triangular mesh \cite{schops2019bad} are storage efficient, however expensive to retrieve map information from. Generative map modeling using GMMs requires only the storage of the means, variances, and mixing proportions. GMM-maps easily adapts to scene complexity, that is, for more complex scenes, we can use more mixture components as necessary.  

\subsection{Adapting Maps to Depth Mispredictions}
In Figure \ref{fig:depthNN}, lightweight depth network with fewer parameters or layers induces significant inaccuracies in the predicted depth map. Therefore, the accuracy of pose estimation suffers. We discuss integrated learning of depth and pose reasoning to overcome such deficiencies of lightweight predictor.

In Figure \ref{fig:mainfig}, we integrate a multi-layer perceptron (MLP)-based learnable transform (size: 3$\times$128$\times$3) to the original point-cloud (PC) map that minimizes the impact of lightweight depth predictor by translating and/or rotating map points adaptively to systematic inaccuracies of the predictor. The last layer of the depth predictor is also tuned. A joint training of map transformations and depth predictor is quite expensive since each update iteration involves nested sampling and particle filtering steps. The complexity of parameter filtering can be significantly minimized using techniques such as hierarchical GMM representations \cite{Dhawale-2020-121381}, beat-tracking \cite{heydari2021don}, \textit{etc.}, however, the resultant formulation is non-differentiable, precluding gradient descent-based optimization.

To circumvent the training complexity, instead of directly minimizing $\ell_2$ norm of the predicted and ground truth pose trajectory, we minimize the negative log-likelihood (NLL) of input image projection via lightweight depth predictor onto the adapted domain maps. Thus, due to the differentiability of the corresponding loss function, the training can be efficiently implemented using standard optimization tools. However, such indirect training of map transforms and depth network is susceptible to overfitting. The loss function focuses on a minimal number of mixture functions in the proximity of ground truth, and it can significantly distort the structural correspondence among the original mixture functions. To alleviate this, we also regularize the loss function by penalizing the distance of the original and adapted map using KL (Kullback Leibler) divergence. Thus, the loss function for the joint training of map transforms, and depth layer is given as:
\begin{equation}
    \mathcal{L}(\theta_\text{M}, \theta_\text{D}) = -\sum_i\text{log}\mathcal{M}_{A,\theta_\text{M}}(\mathcal{D}_{\mathcal{T}_I^i,\mathcal{T}_L^i,\theta_\text{D}}) + \lambda D_\text{KL}(\mathcal{M},\mathcal{M}_A)
\end{equation}
Here, $\theta_\text{M}$ are the parameters for map transformation, and $\theta_\text{D}$ are the parameters of the last layer of depth predictor. For a trajectory $\mathcal{T}$, $\mathcal{T}_I$ represents the set of input images and $\mathcal{T}_L$ corresponding pose labels. $\mathcal{M}$ is the original domain map, and $\mathcal{M}_A$ is the adapted map to compensate for inaccuracies of the lightweight predictor. Both $\mathcal{M}$ and $\mathcal{M}_A$ are represented as GMMs. $\mathcal{D}_{\mathcal{T}_I^i,\mathcal{T}_L^i,\theta_\text{D}}$ is the projection of predicted depth map of trajectory image $\mathcal{T}_I^i$ to 3D space by pin-hole camera model and assuming camera pose at the ground truth label $\mathcal{T}_L^i$. 

\begin{figure}[t]
  \centering
  \includegraphics[width=\textwidth]{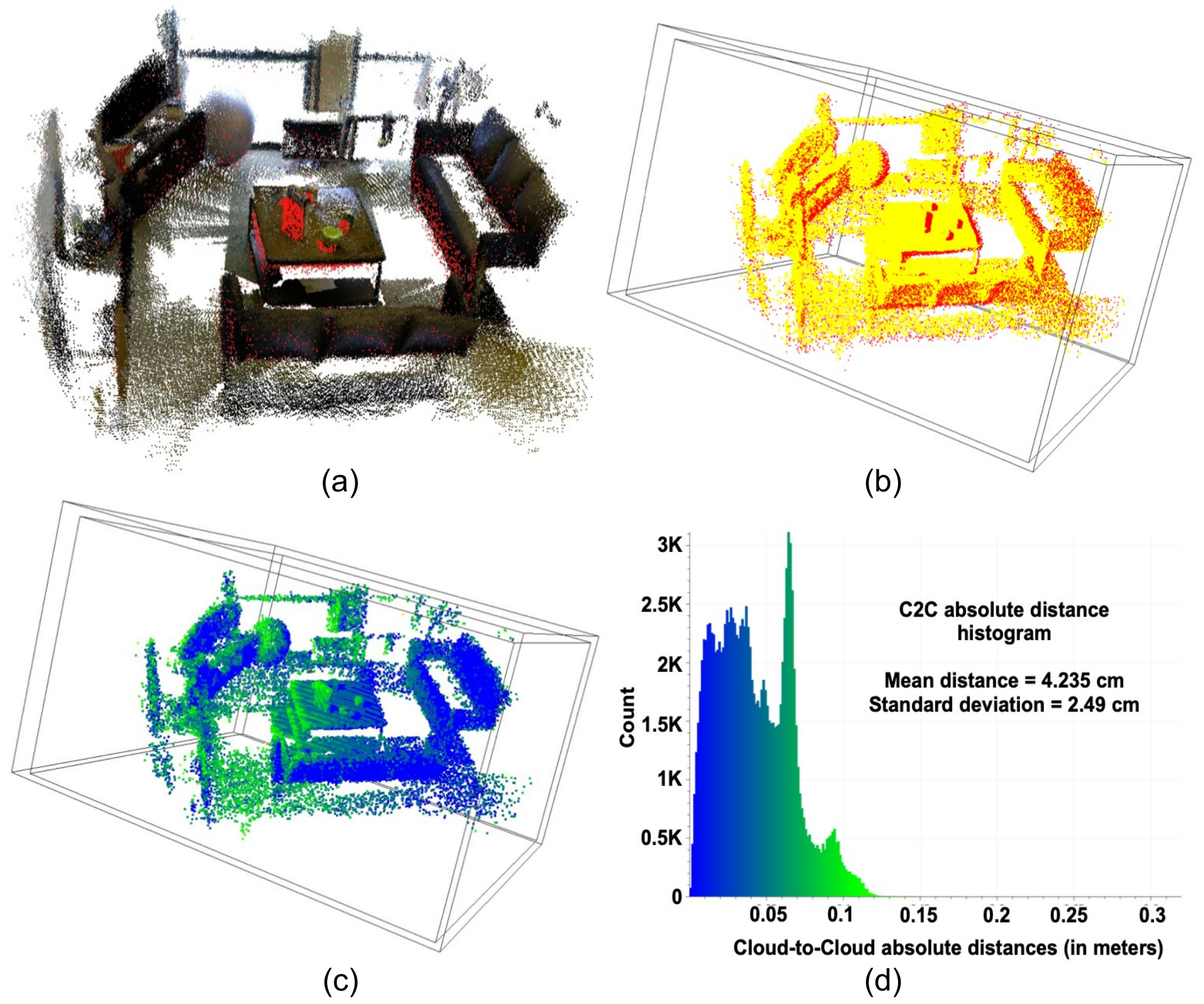}
  \caption{(a) Merged original and transformed point-cloud maps. (b) Bounding boxes of the two maps with the original map in yellow and in red is the transformed map. (c) Relative distance coloring on the reference map data. (d) Histogram of cloud-to-cloud distances.}
  \label{fig:PCtransforms}
\end{figure}

In (1), the regularization term requires computing the KL divergence between the original and adapted maps, namely $\mathcal{M}$ and $\mathcal{M}_A$ respectively. KL divergence of two Gaussian functions is defined in closed form but cannot be analytically extracted for two GMMs. In the proposed framework, original and adapted maps, $\mathcal{M}$ and $\mathcal{M}_A$, have the same number of mixture components, and with a strong enough regularization coefficient ($\lambda$), the relative correspondence among mixture functions maintains, i.e., for i$^\text{th}$ mixture function in $\mathcal{M}$, the nearest mixture function in $\mathcal{M}_A$ has the same index. Leveraging these attributes, the KL divergence of $\mathcal{M}$ and $\mathcal{M}_A$ can be approximated using Goldberger's approximation as \cite{hershey2007approximating}
\begin{equation}
    D_\text{KL}(\mathcal{M},\mathcal{M}_A)  \approx \sum_i \pi_i \Big(D_\text{KL}(M_i,M_{A,i}) + \log \frac{\pi_i}{\pi_{A,i}}\Big) 
\end{equation}
Here, $M_i$ is the i$^\text{th}$ mixture component of $\mathcal{M}$, and $M_{A,i}$ is the corresponding component in $\mathcal{M}_A$. $\pi_i$ is $M_i$'s weight and $\pi_{A,i}$ is $M_{A,i}$'s weight. The KL divergence of $M_i$ and $M_{A,i}$, i.e., $D_\text{KL}(M_i,M_{A,i})$ is analytically defined. Thus, $D_\text{KL}(\mathcal{M},\mathcal{M}_A)$ can be efficiently computed and is differentiable. 

Figure \ref{fig:PCtransforms} shows the point cloud adaptations of Scene-02 in RGBD dataset \cite{lai2014unsupervised} using the method. Figure \ref{fig:PCtransforms}(a) contains both the original and adapted point-cloud (PC) maps. In Figure \ref{fig:PCtransforms}(b), the reference or original map's 3D points are in yellow while the adapted PC is in red to highlight the adaptation difference. In Figure \ref{fig:PCtransforms}(c), the reference point cloud's 3D points are color-coded based on the relative distance of corresponding points in the adapted map. The cloud-to-cloud (C2C) distance histogram is shown in Figure \ref{fig:PCtransforms}(d). Thus, the results demonstrate that only a minimal tweaking of map data is sufficient to improve pose accuracy (evident in later results) despite extremely lightweight depth prediction.

\begin{figure}[t]
  \centering
  \includegraphics[width=0.9\columnwidth]{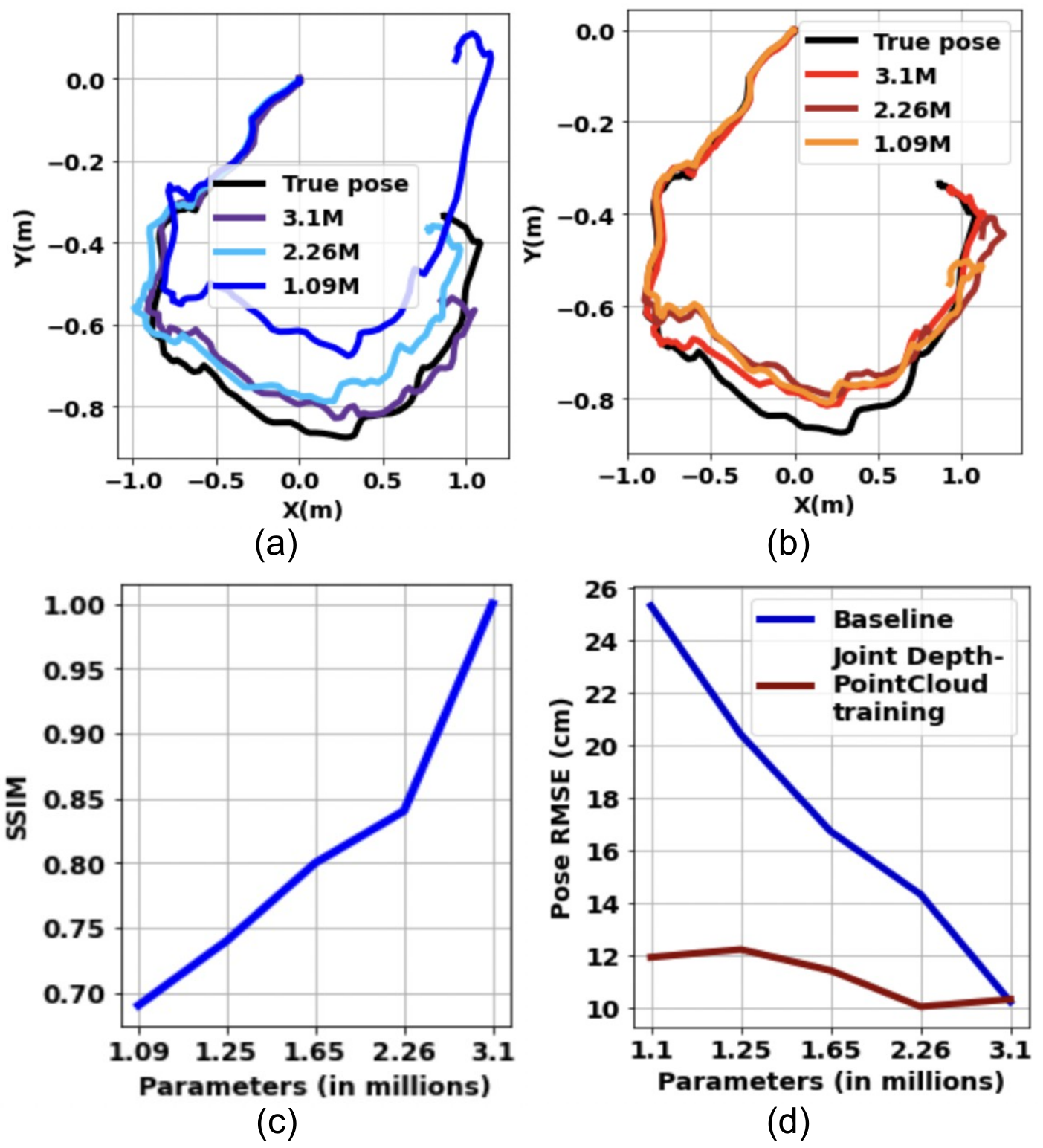}
  \caption{(a) Predicted pose trajectory using the proposed integrated depth estimator and pose filter for varying depth network sizes without joint training of depth estimator and pose filter. (b) Pose predictions for varying depth network sizes using the proposed technique with jointly training learnable map transforms and lightweight depth predictor. (c) The structural similarity index of depth predictions reduces for a reduction in network size. (d) Comparison of Pose errors (RMSEs) for baseline and proposed technique.} 
  \label{fig:paramvarplots}
\end{figure}
\vspace{-1em}

\section{Results and Discussion}
Figures \ref{fig:paramvarplots}(a) and (b) compare the predicted pose trajectory (for varying depth network size) from the proposed monocular localization against an equivalent framework where joint training of depth network and filtering model is not performed. The comparison uses the RGBD scenes dataset \cite{lai2014unsupervised}. Figure \ref{fig:paramvarplots}(c) shows the corresponding degradation in depth images, measured using structural similarity index measure (SSIM). In Figure \ref{fig:paramvarplots}(d), despite significant degradation in depth image quality and reduction of depth predictor to one-third parameters, the proposed joint training maintains pose prediction accuracy by learning and adapting against systematic inaccuracies in depth prediction. Another crucial feature is that the original depth predictor can be trained on any dataset, and then tuned (on the last layer) for the application domain. For example, in the presented results, the original depth network was trained on NYU-Depth \cite{silberman2012indoor} and applied on RGBD scenes \cite{lai2014unsupervised}. Thus, the predictor has access to vast training data that can be independent of application domain.      

\begin{figure}[t]
  \centering
  \includegraphics[width=0.95\columnwidth]{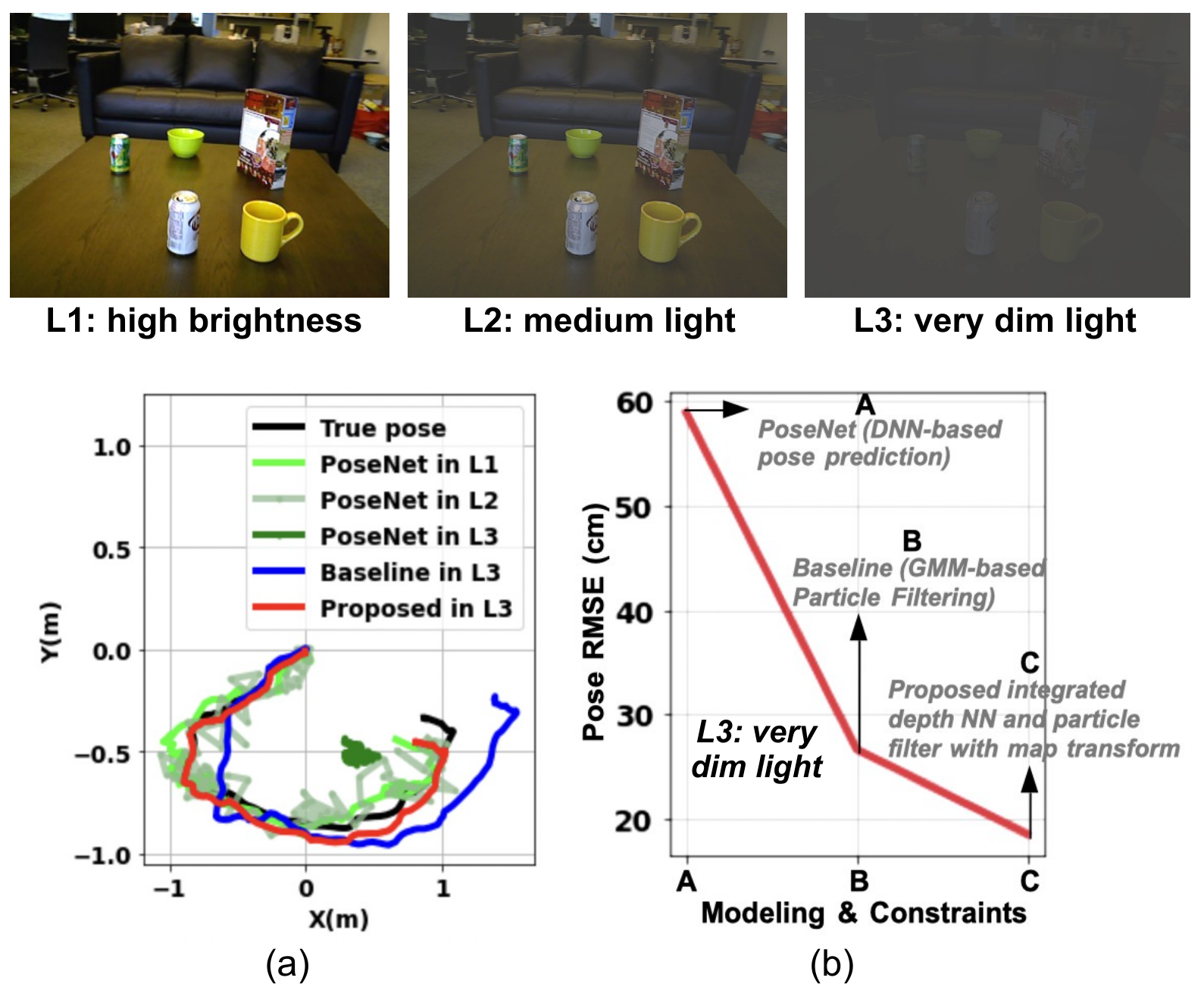}
  \caption{On top: Indoor RGB image captured in different lighting conditions. (a) Comparison of pose trajectories for MobileNetv2-based PoseNet, baseline GMM map-based pose filtering, and proposed integrated framework of depth estimator and pose filter in various lighting conditions. (c) Pose error (RMSE) plot in very dim light for various models.} 
  \label{fig:lightvarplots}
\end{figure}

Figure \ref{fig:lightvarplots} demonstrates the resilience of proposed crossmodal pose prediction against DNN by considering extreme lighting variations. An equivalent MobileNet-v2-based PoseNet \cite{kendall2015posenet} is utilized as DNN for the comparisons. On the top, input images are subjected to extreme lighting variations using models in \cite{lai2014unsupervised} (L1: high brightness, L2: medium light, and L3: very dim light). Figure \ref{fig:lightvarplots}(a) compares trajectories from PoseNet and our framework (with and without the joint training). In all cases,  equivalent sized models are considered, shown in Table I. In Figure \ref{fig:lightvarplots}(b), our framework is significantly more accurate than PoseNet in very dim light (L3) conditions due to in-built filtering loops, demonstrating superiority of crossmodal estimates than DNN-only estimates. 

\vspace{-2em}
\begin{figure}[t]
  \centering
  \includegraphics[width=\columnwidth]{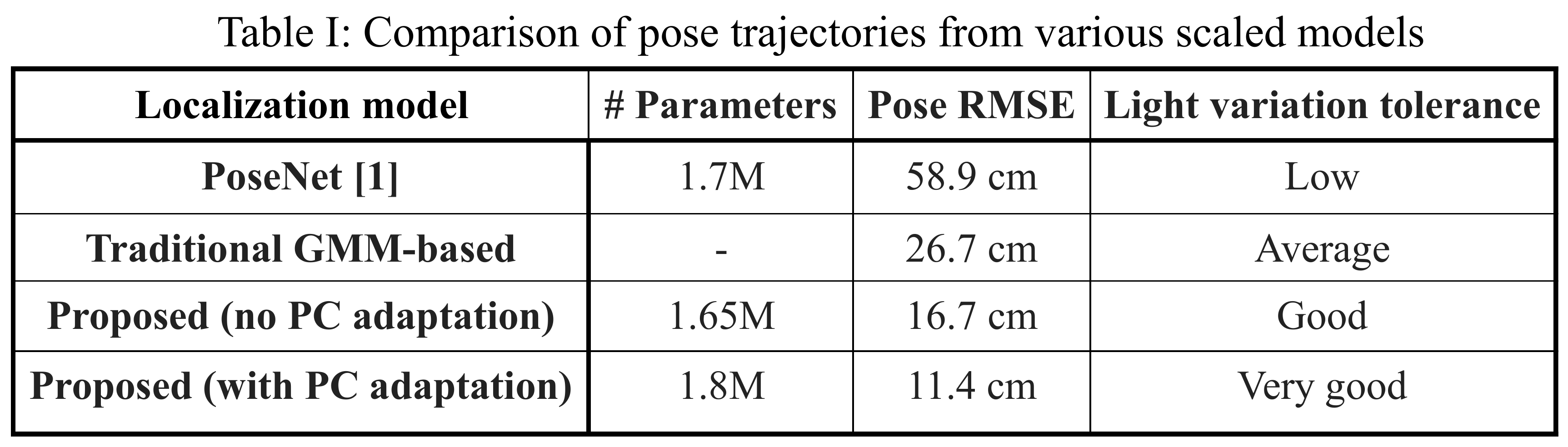}
  \label{fig:table}
\end{figure}

\section{Conclusions}
We presented a novel monocular localization framework by jointly learning depth estimates and map transforms. Compared to standard DNNs for pose estimates, such as PoseNet, the proposed approach is significantly more tolerant to model size scalability and environmental variations. Open representation of map and depth estimates in our approach also allows faster updates and resource efficiency by availing intermediate feature maps for other automation objectives, such as depth maps for obstacle avoidance.


\bibliographystyle{IEEEtran}
\bibliography{main.bib}
\end{document}